\documentclass[runningheads]{llncs}

\usepackage{eccv}

\usepackage{eccvabbrv}

\usepackage{graphicx}
\usepackage{booktabs}

\usepackage[accsupp]{axessibility}  

\usepackage{amsmath}
\usepackage{amssymb}
\usepackage{float, amsfonts, amsmath, amssymb, multirow, verbatim,array, bm}
\usepackage{threeparttable}
\usepackage{algorithm, algpseudocode} 
\usepackage{xcolor}
\usepackage{caption}
\usepackage{tabularx}
\usepackage{microtype}
\usepackage{multirow, makecell}

\usepackage{xassoccnt}
\newcounter{realpage}
\DeclareAssociatedCounters{page}{realpage}
\AtBeginDocument{%
  \stepcounter{realpage}
}

\newcommand{\safesim}{\textsc{Safe-Sim}\xspace}

\usepackage[accsupp]{axessibility}  

\usepackage{hyperref}

\usepackage{orcidlink}

\usepackage[accsupp]{axessibility}  

\begin{document}

\title{\safesim: Safety-Critical Closed-Loop Traffic Simulation with Diffusion-Controllable  Adversaries}

\titlerunning{Safety-Critical Closed-Loop Traffic
Simulation with Controllable Adversaries}


\titlerunning{Safety-Critical Closed-Loop Traffic Simulation with Controllable Adversaries}



\author{ Wei-Jer Chang\inst{1} \and
Francesco Pittaluga\inst{2} \and
Masayoshi Tomizuka\inst{1} \and
Wei Zhan\inst{1} \and
Manmohan Chandraker\inst{2,3}
}


\authorrunning{W.J. Chang et al.}

\institute{$^1$ UC Berkeley \hspace{0.5cm} $^2$ NEC Labs America \hspace{0.5cm} $^3$ UC San Diego}

\maketitle

\begin{abstract}
Evaluating the performance of autonomous vehicle planning algorithms necessitates simulating long-tail safety-critical traffic scenarios. However, traditional methods for generating such scenarios often fall short in terms of controllability and realism; they also neglect the dynamics of agent interactions. To address these limitations, we introduce \safesim, a novel diffusion-based controllable closed-loop safety-critical simulation framework. Our approach yields two distinct advantages: 1) generating realistic long-tail safety-critical scenarios that closely reflect real-world conditions, and 2) providing controllable adversarial behavior for more comprehensive and interactive evaluations. We develop a novel approach to simulate safety-critical scenarios through an adversarial term in the denoising process of diffusion models, which allows an adversarial agent to challenge a planner with plausible maneuvers while all agents in the scene exhibit reactive and realistic behaviors. Furthermore, we propose novel guidance objectives and a partial diffusion process that enables users to control key aspects of the scenarios, such as the collision type and aggressiveness of the adversarial agent, while maintaining the realism of the behavior. We validate our framework empirically using the nuScenes and nuPlan datasets across multiple planners, demonstrating improvements in both realism and controllability. These findings affirm that diffusion models provide a robust and versatile foundation for safety-critical, interactive traffic simulation, extending their utility across the broader autonomous driving landscape. Project website: \href{https://safe-sim.github.io/}{https://safe-sim.github.io/}.
\end{abstract}

\section{Introduction}

\label{sec:intro}
A key safety feature of autonomous vehicles (AVs) is their ability to navigate near-collision events in real-world scenarios. However, these events rarely occur on roads and testing AVs in such high-risk situations on public roads is unsafe. Therefore, simulation is indispensable in the development and assessment of AVs, providing a safe and reliable means to study their safety and dependability. A critical aspect of simulation is modeling the behavior of other road users, since AVs must learn to interact with them safely. 

A common method of safety-critical testing of AVs involves manually designing scenarios that could potentially lead to failures, such as collisions. While this approach allows for targeted testing, it is inherently limited in scalability and lacks the comprehensiveness required for thorough evaluation \cite{wang2021advsim, ding2020learning, multimodal_safety_critical}. Some recent works focus on automatically generating challenging scenarios that cause planners to fail, but their emphasis has been mostly on static scenario generation rather than dynamic, \textit{closed-loop} simulations. This results in a critical gap: the behavior of other agents often does not adapt or respond to the planner's actions, which is essential for a comprehensive safety evaluation. Furthermore, the results from these simulations often lack \textit{controllability}, typically producing only a single adversarial outcome per scenario without the flexibility to explore a range of conditions and responses.  

\setlength\tabcolsep{1.5 pt}
\begin{table}[!t]
\centering
\begin{tabular}{lcccccc}
\toprule
\textbf{Method} & \begin{tabular}[c]{@{}c@{}}\textbf{Safety-}\\\textbf{Critical}\end{tabular} & \begin{tabular}[c]{@{}c@{}}\textbf{Controllable}\end{tabular} & \begin{tabular}[c]{@{}c@{}}\textbf{Controllable}\\\textbf{Adversary}\end{tabular} & \begin{tabular}[c]{@{}c@{}}\textbf{Evaluate}\\\textbf{Planner}\end{tabular} & \begin{tabular}[c]{@{}c@{}}\textbf{Closed-}\\\textbf{Loop}\end{tabular} & \begin{tabular}[c]{@{}c@{}}\textbf{Real-}\\\textbf{World}\end{tabular} \\
\midrule
CTG \cite{CTG}            & \textcolor{red}{$\mathbf{\times}$} & \textbf{\textcolor{Green}{\checkmark}} & \textcolor{red}{$\mathbf{\times}$} & \textcolor{red}{$\mathbf{\times}$} & \textbf{\textcolor{Green}{\checkmark}} & \textbf{\textcolor{Green}{\checkmark}} \\
CTG++ \cite{ctg++}            & \textcolor{red}{$\mathbf{\times}$} & \textbf{\textcolor{Green}{\checkmark}} & \textcolor{red}{$\mathbf{\times}$} & \textcolor{red}{$\mathbf{\times}$} & \textbf{\textcolor{Green}{\checkmark}} & \textbf{\textcolor{Green}{\checkmark}} \\
STRIVE \cite{strive} & \textbf{\textcolor{Green}{\checkmark}} & \textcolor{red}{$\mathbf{\times}$} & \textcolor{red}{$\mathbf{\times}$} & \textbf{\textcolor{Green}{\checkmark}} & \textbf{\textcolor{Green}{\checkmark}} & \textbf{\textcolor{Green}{\checkmark}} \\
DiffScene \cite{xu2023diffscene}       & \textbf{\textcolor{Green}{\checkmark}} & \textbf{\textcolor{Green}{\checkmark}} & \textcolor{red}{$\mathbf{\times}$} & \textbf{\textcolor{Green}{\checkmark}} & \textcolor{red}{$\mathbf{\times}$} & \textcolor{red}{$\mathbf{\times}$} \\
\safesim (Ours)            & \textbf{\textcolor{Green}{\checkmark}} & \textbf{\textcolor{Green}{\checkmark}} & \textbf{\textcolor{Green}{\checkmark}} & \textbf{\textcolor{Green}{\checkmark}} & \textbf{\textcolor{Green}{\checkmark}} & \textbf{\textcolor{Green}{\checkmark}} \\
\bottomrule
\end{tabular}
\vspace{.2em}
\caption{\textbf{Comparison of methods.} Our contribution is the development of a framework for (a) safety-critical (b) closed-loop (c) controllable adversarial simulations. These aspects are not concurrently present in previous frameworks. We formulate a novel partial diffusion with novel guidance functions for stable long-term simulation and are the first to enable an ego planner to be tested against controllable adversaries with varied behavior patterns. \vspace{-3em}}
\label{tab:compare}
\end{table}

\setlength\tabcolsep{6 pt}

\begin{figure}[!t]
    \centering
    \includegraphics[width=\linewidth]{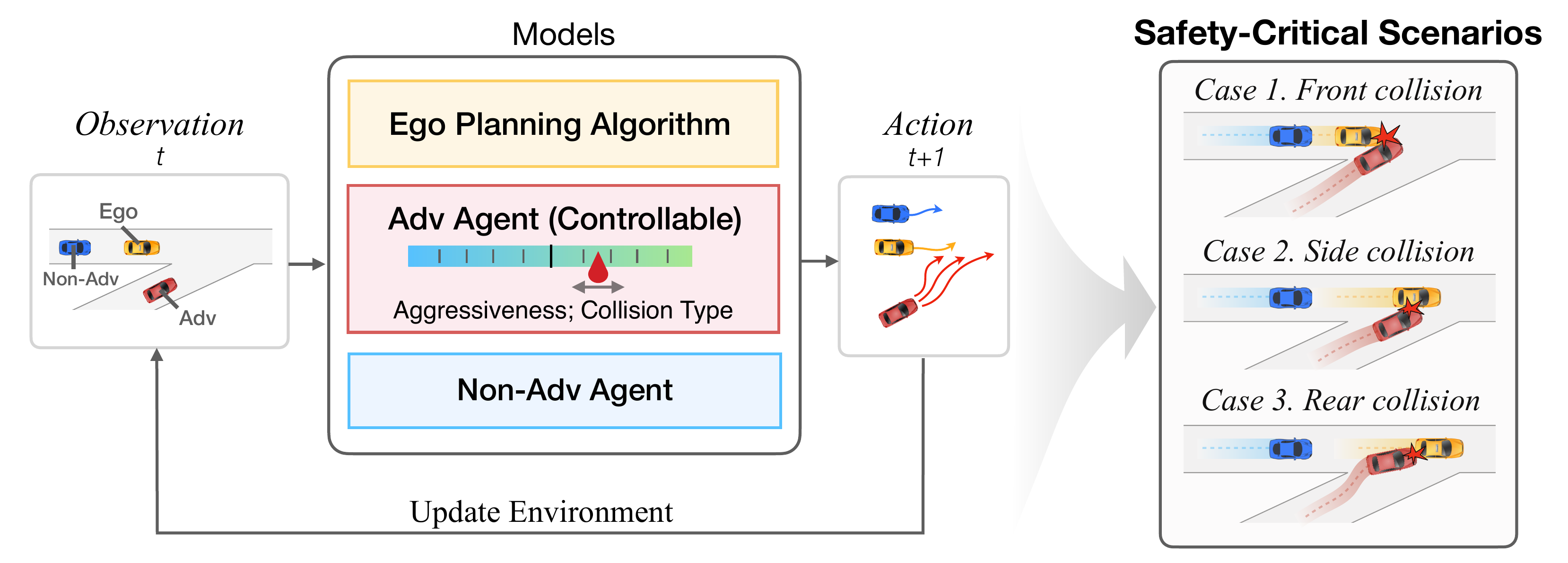}
    \caption{\textbf{Overview of \safesim Framework for Controllable Safety-Critical Closed-Loop Simulation.} This framework evaluates a planner within scenarios featuring multiple controllable reactive agents. These agents have two distinct roles: adversarial agents, which actively challenge the planner by exhibiting controllable adversarial behaviors such as specific collision types and levels of aggressiveness, and non-adversarial agents, which follow normal driving behavior to maintain the realism of the entire scene. Such a setup facilitates the generation of various realistic, interactive, and safety-critical scenarios, providing a thorough evaluation of the planner’s capabilities.}
    \label{fig:overview}
\end{figure}

In this work, we introduce \safesim, a {\em closed-loop} simulation framework for generating safety-critical scenarios, with a particular emphasis on {\em controllability} and {\em realism} for the behavior of agents, which allows simulations over a long-horizon as needed to evaluate AV planning algorithms (Fig.~\ref{fig:overview}). Different from prior works \cite{CTG, ctg++, strive, xu2023diffscene} that primarily adhere to rule-constraint satisfaction, our approach enhances \textit{controllability} by modulating adversarial vehicle behaviors within identical scenarios, thereby facilitating a broader exploration of potential outcomes. See \cref{tab:compare} for a comprehensive comparison of these approaches.

Our approach builds upon recent developments in controllable diffusion models \cite{diffuser, CTG, trace}. Specifically, we adopt a test-time guidance to direct the denoising phase of the diffusion process, using the gradients from differentiable objectives to enhance scenario generation, enabling generation of adversarial scenarios in which and adversarial agent collides with the ego agent behaving according to specific planning policy. Additionally, we develop an novel approach, which we refer to as Partial Diffusion that introduces trajectory proposals into the diffusion process to provide a high degree of controllability over the type of collision scenario. Overall, our balanced integration of adversarial objectives with regularization during the guidance phase combined with Partial Diffusion allows for refined control over the conditions of the generated scenarios, ensuring both their realism and relevance to safety-critical testing.

In our study, we use the nuScenes \cite{caesar2020nuscenes} and nuPlan \cite{caesar2021nuplan} datasets to evaluate the efficacy of our method in generating safety-critical closed-loop simulations. Our results demonstrate a marked improvement in the controllability and realism of scenarios compared to previous adversarial scenario generation methods. Furthermore, we showcase the advantage of our proposed framework in varying the safety-criticality and collision types of scenarios. These attributes make our approach particularly well-suited for the closed-loop simulation of AVs, providing a more reliable and comprehensive framework for safety evaluation.

\section{Related Work}
\label{sec:related}

\subsection{Traffic Simulation}
Traffic simulation can broadly be categorized into two main groups: heuristic-based and learning-based methods. In heuristic-based methods, agents are controlled by human-specified rules, such as the Intelligent Driver Model (IDM) \cite{treiber2000congested} to follow a leading vehicle while maintaining a safe following distance. However, these methods have modeling capacity issues and may not reflect the real traffic distribution; a large domain gap limits its usage for planning evaluation. To close the gap, data-driven approaches learn from real driving datasets to imitate real-world behavior \cite{suo2021trafficsim,BITS,suo2023mixsim, philion2024trajeglish}. TrafficSim \cite{suo2021trafficsim} utilizes a trained variational autoencoder for scene-level traffic simulation, while BITS \cite{BITS} combines high-level goal inference with low-level driving behavior imitation to enhance the realism of simulated driving behavior. Recently, the Waymo SimAgents challenge has focused on whether simulators can accurately represent real-world driving distributions \cite{waymo_sim_agent}. However, little work has specifically focused on behavior simulation for generating safety-critical, long-tail scenarios.

Diffusion models \cite{Midjourney,Dall-e-2,rombach2022high} have shown significant promise for synthetic image generation. One of the key advantages of diffusion models is controllability, which can take the form of classifier \cite{dhariwal2021diffusion}, classifier-free \cite{ho2022classifier}, and reconstruction \cite{ho2022imagen} guidance. Recently, controllable diffusion models have been employed for planning and traffic simulation \cite{CTG,trace,ctg++} via guidance. We adopt trajectory diffusion models to develop a novel approach for generate safety-critical realistic traffic simulations in which an adversarial agent collides with an ego planner agent. Other works \cite{scenecontrol2024_waabi, pronovost2023scenario_zoox} use diffusion with guidance or conditioning to achieve controllable scene initialization. In contrast, we emphasize closed-loop, controllable adversarial behavior simulation based on real-world data initialization rather than scene initialization.

\subsection{Safety-Critical Traffic Simulation}
Safety-critical traffic generation plays a crucial role in training and evaluating AV systems, enhancing their capability to navigate diverse real-world scenarios and enhancing robustness. Gradient-based methods that leverage back-propagation to create safety-critical scenarios have been proposed to evaluate AV prediction and planning models\cite{hanselmann2022king, cao2022advdo}. Hanselmann et al. use kinematic gradients to modify vehicle trajectories, with the goal of improving the robustness of imitation learning planners \cite{hanselmann2022king}.  Cao et al. have developed a model with differentiable dynamics, enabling the generation of realistic adversarial trajectories for trajectory prediction models through backpropagation techniques \cite{cao2022advdo}.
Black-box optimization approaches include perturbing actions based on kinematic bicycle models \cite{wang2021advsim} and using Bayesian Optimization to create adversarial self-driving scenarios that escalate collision risks with simulated entities \cite{abeysirigoonawardena2019generating}. Zhang et al. target trajectory prediction models via white- and black-box attacks that adversarially perturb real driving trajectories \cite{zhang2022adversarial}. For an extensive review of this topic, we direct readers to Ding et al. \cite{ding2023survey}.

The field has recently seen advancements in data-driven methods for safety-critical scenario generation \cite{yin2021diverse, xu2023diffscene, strive}. For instance, Xu et al. introduce a diffusion-based approach in CARLA, applying various adversarial optimization objectives to guide the diffusion process for safety-critical scenario generation \cite{xu2023diffscene}. Rempe et al. proposed STRIVE to utilize gradient-based adversarial optimization on the latent space, constrained by a graph-based CVAE traffic motion model, to generate realistic safety-critical scenarios for rule-based planners \cite{strive}. However, a common limitation in these approaches is the absence of closed-loop interaction, essential for accurately simulating interactive real-world driving.

\section{Problem Formulation}
\label{subsec:problem}

We consider a simulated interactive traffic scenario consisting of $N$ agents; one is the ego vehicle controlled by the \textit{planner} $\pi$, and the remaining $N-1$ are reactive agents modeled by a function $g$. Our objective is to create a safety-critical \textit{closed-loop} collision simulation, where reactive agents demonstrate \textit{realistic}, \textit{controllable} behavior. Of the $N-1$ reactive agents, one or a subset is considered the adversarial agents (denoted as agent $a$), meant to collide with the ego vehicle.

\textcolor{black}{The adversarial agent, formulated within the reactive agent model $g$, is governed by an adversarial term designed (detailed in \cref{sec:guided_advsec}) to be both controllable and adversarial to the planner $\pi$.  This setup allows the adversarial agent to pose direct challenges to $\pi$, testing its resilience in complex scenarios. Concurrently, the other non-adversarial agents, also controlled by $g$ with varying parameters, emulate authentic, reactive behaviors, thus enriching the simulation scenario with realistic and diverse traffic conditions. The dual role of the adversarial agent and non-adversarial agents ensures that while it challenges $\pi$, the overall simulation environment plausibly represents real-world driving conditions.}

At any given timestep $t$, the states of the $N$ vehicles are represented as $\mathbf{s}_t = [\mathbf{s}^1_t, \ldots, \mathbf{s}^N_t]$, where $\mathbf{s}^i_t = (x^i_t, y^i_t, v^i_t, \theta^i_t)$ indicates the 2D position, speed, and yaw of vehicle $i$. The corresponding actions for each vehicle are $\mathbf{a}_t = [\mathbf{a}^1_t, \ldots, \mathbf{a}^N_t]$, with $\mathbf{a}^i_t = (\dot{v}^i_t, \dot{\theta}^i_t)$ representing the acceleration and yaw rate. To predict the state at the next timestep $t+1$, a transition function $f$ is used, which computes $\mathbf{s}_{t+1} = f(\mathbf{s}_t, \mathbf{a}_t)$ based on current state and action. We adopt unicycle dynamics as the transition function.

Each agent's decision context is $\mathbf{c}^i_t$, which includes the agent-centric map $I^i$ and the $T_{\text{hist}}$ historical states of neighboring vehicles from time $t-T_{\text{hist}}$ to $t$, defined as $\mathbf{s}_{t-T_{\text{hist}}:t} = \{\mathbf{s}_{t-T_{\text{hist}}}, \ldots, \mathbf{s}_t\}$. In closed-loop traffic simulation, each agent continuously generates and updates its trajectory based on the current decision context $\mathbf{c}^i_t$. After generating a trajectory, the simulation executes the first few steps of the planned actions before updating $\mathbf{c}^i_t$ and re-planning. See \cref{sec:exp_implem} for more implementation details.

\paragraph{Planner $\pi$}
The planner $\pi$ determines the ego vehicle's future trajectory over a time horizon $t$ to $t+T$. The planned state sequence is denoted by $s^1_{t:t+T} = \pi(\mathbf{c}^1_t)$, where $\pi(\mathbf{c}^1_t)$ processes the historical states and map data within $\mathbf{c}^1_t$ to plan future states based on the current scene context.

\paragraph{Reactive Agents $g$}
The reactive agent model $g$, parameterized by $\theta$, is designed to simulate the behavior of the $N-1$ non-ego vehicles, represented by the set $\{s^i_{t:t+T}\}_{i=2}^{N}$. Each vehicle's state sequence, $s^i_{t:t+T}$, is generated by $g_{\theta}(\mathbf{c}^i_t, \psi_i)$, which incorporates the decision context $\mathbf{c}^i_t$ and a set of control parameters $\psi_i$ unique to each agent. These parameters $\psi_i$ enable the fine-tuning of individual behaviors within the simulation.
In our approach, we train the model $g$ on real-world driving data to ensure the trajectories it produces are not only controllable, supporting the generation of various safety-critical scenarios, but also realistic.

\section{Diffusion Models for Traffic Simulation}\label{sec:traj_diff}

For closed-loop safety-critical traffic simulation, the reactive agents, especially the adversarial agent, should be 1) controllable, and 2) realistic. With recent advances in controllable diffusion models \cite{diffuser,CTG,trace}, we adopt trajectory diffusion models to generate realistic simulations. 

We define the model's operational trajectory as $\tau$, which comprises both action and state sequences: $\tau:= [\tau_a, \tau_s]$. Specifically, $\tau_a := [a_0, \ldots, a_{T-1}]$ represents the sequence of actions, while $\tau_s := [s_1, \ldots, s_T]$ denotes the corresponding sequence of states. Following the approach described in \cite{CTG}, our model predicts the action sequence $\tau_a$, and the state sequence $\tau_s$ can be derived starting from the initial state $s_0$ and dynamic model $f$. 

A diffusion model generates a trajectory by reversing a process that incrementally adds noise. Starting with an actual trajectory $\tau_0$ sampled from the data distribution $q(\tau_0)$, a sequence of increasingly noisy trajectories $ (\tau_1, \tau_2, \ldots, \tau_K )$ is produced via a forward noising process. Each trajectory \( \tau_k \) at step \( k \) is generated by adding Gaussian noise parameterized by a predefined variance schedule \( \beta_k \) \cite{ddpm}:
\begin{equation}
q(\tau_{1:K} | \tau_0) := \prod_{k=1}^{K} q(\tau_k | \tau_{k-1}),
\end{equation}
\begin{equation}
q(\tau_k | \tau_{k-1}) := \mathcal{N}(\tau_k; \sqrt{1 - \beta_k}\tau_{k-1}, \beta_k\mathbf{I}).
\end{equation}
The noising process gradually obscures the data, where the final noisy version  \( q(\tau_K) \) approaches \( \mathcal{N}(\tau_K; \mathbf{0}, \mathbf{I}) \).
The trajectory generation process is then achieved by learning the reverse of this noising process. Given a noisy trajectory \( \tau_K \), the model learns to denoise it back to \( \tau_0 \) through a sequence of reverse steps. Each reverse step is modeled as:
\begin{equation}\label{eq:3}
p_{\theta}(\tau_{k-1} | \tau_k, \mathbf{c}) := \mathcal{N}(\tau_{k-1}; \mu_\theta(\tau_k, k, \mathbf{c}), \Sigma_k),
\end{equation}
where $\theta$ are learned functions that predict the mean \( \mu \) of the reverse step, and $\Sigma_k$ is a fixed schedule. By iteratively applying the reverse process, the model learns a trajectory distribution, effectively generating a plausible future trajectory from a noisy start.

During the trajectory prediction phase, the model estimates the final clean trajectory denoted by \( \hat{\tau}_0 \). This estimated trajectory is used to compute the mean \( \mu \) as described in \cite{diffusion_clean}. For more details, see supplementary material \cref{sec:supp_implmentation_details}.

\section{Diffusion Models for Safety-Critical Traffic Simulation}\label{sec:guided_advsec}


The diffusion model, once trained on realistic trajectory data, inherently reflects the behavioral patterns present in its training distribution. However, to effectively simulate and analyze safety-critical scenarios, there is a crucial need for a mechanism that allows for the controlled manipulation of agent behaviors \cite{diffuser,CTG}. This is particularly important for generating adversarial behaviors and ensuring scene consistency in simulations. 


\begin{figure}[!!t]
    \centering
    \includegraphics[width=\linewidth]{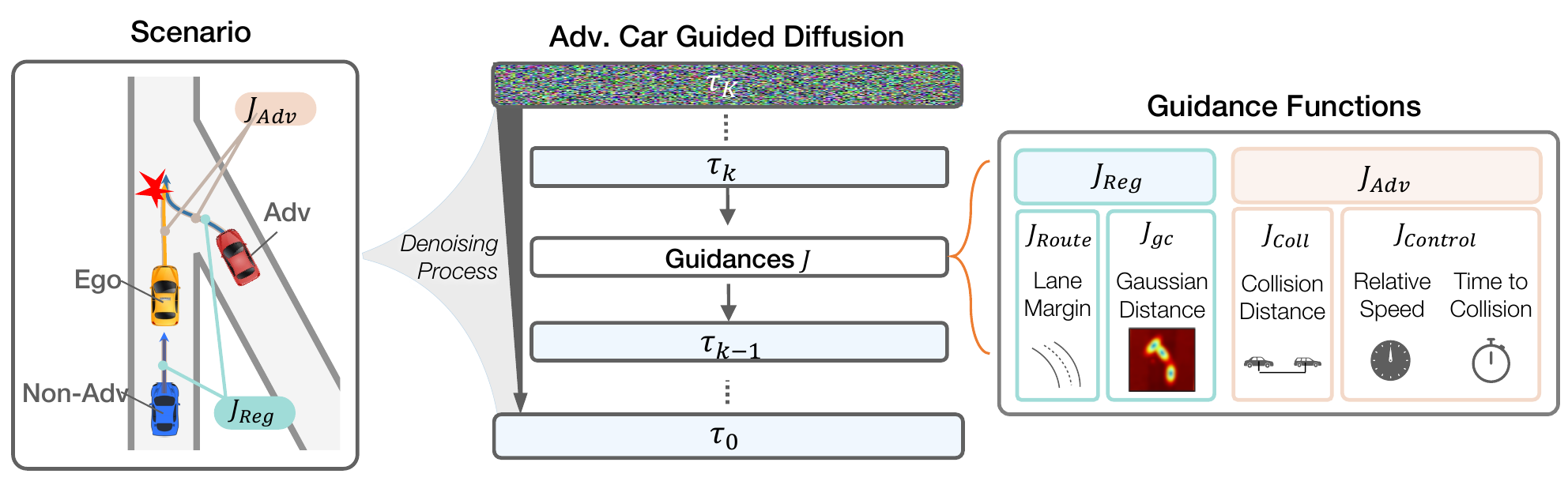}
    \caption{\textbf{Guided Diffusion Process for the Adversarial Agent.} This process optimizes the adversarial agent's trajectory using the adversarial cost function $J_{\text{adv}}$ to the ego vehicle. In particular, we introduce $J_{\text{control}}$ to vary the adversarial behavior. Simultaneously, it applies regularization through $J_{\text{reg}}$ for maintaining realism.}
    \label{fig:guided_adv}
   
\end{figure}

\subsection{Guiding Reactive Agents}
Our approach specifically introduces guidance to the sampled trajectories at each denoising step, aligning them with predefined objectives \( J(\tau) \). The concept of guidance involves using the gradient of \( J \) to subtly perturb the predicted mean of the model at each denoising step. This process enables the generation of trajectories that not only reflect realistic behavior but also cater to specific simulation needs, such as adversarial testing and maintaining scene consistency over extended periods. We adopt the reconstruction guidance (clean guidance) introduced from \cite{video_diffusion, trace}:
\begin{equation}
\tilde{\tau}_0 = \hat{\tau}_0 - \alpha \Sigma_k \nabla_{\tau_k} J(\hat{\tau}_0)
\end{equation}
This strategy improves guidance robustness, yielding
smoother, more stable trajectories without the usual numerical issues from noisy data.

\textcolor{black}{In practice, diversifying the behavior of adversarial agents within the \textit{same} scenarios is crucial for a thorough assessment of AVs. Despite the significance of this challenge, it remains largely unexplored in previous works \cite{strive,CTG} }

\textcolor{black}{
The loss function for the non-reactive agents, $J(\tau)$, consists of a collision term $J_{\text{coll}}$, which encourages collisions between the adversarial agent and the ego agent, two control terms $J_{\text{v}}$ and $J_{\text{ttc}}$, which control the relative speed and time-to-collision between the ego and adversarial agent respectively, a regularization term $J_{\text{Gauss}}$, which discourages collisions between the reactive agents, and a route guidance term $J_{\text{route}}$, which discourages the reactive agents from going outside the road: 
\begin{equation}
    J(\tau) = \rho\underbrace{(J_{\text{coll}} + J_{\text{v}} + J_{\text{ttc}})}_{J_{\text{adv}}(\tau)} + \underbrace{J_{\text{route}} + J_{\text{Gauss}}}_{J_{\text{reg}}(\tau)},
\end{equation}
where $\rho$ denotes a scalar weight that determines whether a reactive agent behaves adversarially towards the ego agent, i.e., whether it attempts to collide with the ego agent.
}

\subsubsection{Collision with Planner}
We define $J_{\text{coll}}$ to encourage the collision between the adversarial agent and the ego agent, given by:
\begin{equation}
J_{\text{coll}} = - \sum_{t=1}^{T} d(t) ,
\end{equation}
where $d(t)$ represents the distance between the ego and the adversarial agent at each time step of the planning horizon $T$. The adversarial agent is either pre-selected based on lane proximity or dynamically selected based on the distance to the ego agent. Details are in the supplementary \cref{supp:selecting_adv}.

\subsubsection{Safety Criticality of Collisions}
We control the relative speed $J_v$ between the ego and adversary at each time step ($v^1_t$ and $v^a_t$), and the time-to-collision (TTC) cost $J_{\text{ttc}}$ \cite{nishimura2023rap} to control the safety criticality of potential collisions, with the latter given by:
\begin{equation}
J_{\text{ttc}} = \sum_{t=1}^{T} - \exp\left( -\frac{\tilde{t}_{\text{col}(t)}^2}{2\lambda_t} - \frac{\tilde{d}_{\text{col}(t)}^2}{2\lambda_d} \right),
\end{equation}
where \( \tilde{t}_{\text{col}(t)} \) is the time to collision at time \( t \), \( \tilde{d}_{\text{col}(t)} \) is the distance to collision  and \( \lambda_t \) and \( \lambda_d \) are bandwidth parameters for time and distance. This formula uses a constant velocity assumption. Intuitively, the time-to-collision cost favors scenarios with high relative speeds and challenging collision angles for the ego vehicle to avoid. For details of $J_v$ and $J_{\text{ttc}}$, see supplementary \cref{supp:adv_behavior_def}.




\subsubsection{Route Guidance}
Given an agent's trajectory \( \tau \) and the corresponding route \( r \)—the predefined path on a lane graph from its starting point to its destination—we compute the normal distance of each point \( \tau_t \) on the trajectory to the route at each timestep. We then penalize deviations from the route that exceed a predefined margin \( d \). This process is captured by the following route guidance cost function:

\begin{equation}
J_{\text{route}}(\tau, r) = \sum_{t=1}^{T} \max(0, |d_n(\tau_t, r) - d_m|),
\end{equation}
where \( d_n(\tau_n, r) \) denotes the normal distance from the point \( \tau_t \) on the trajectory to the nearest point on the route \( r \) at timestep \( t \), and \( d_m \) represents the acceptable deviation margin from the route. In contrast to the off-road loss in prior studies \cite{CTG}, our proposed route guidance system more effectively indicates each agent’s intended path, improving adherence to traffic rules as demonstrated in \cref{sec:experiments}. The flexibility of route guidance supports diverse agent interactions, such as modifying routes to encourage lane changes among reactive agents \cite{suo2023mixsim}.

\subsubsection{Gaussian Collision Guidance} 
Given the trajectories of agents, we calculate the Gaussian distance for each pair of agents $(i, j)$ at each timestep $t$ from $1$ to $T$. The Gaussian distance between the agents takes into account both the tangential ($d_{t}$) and normal ($d_{n}$) components of the projected distances. The aggregated Gaussian distance is:
\begin{equation}
J_{\text{Gauss}} = \sum_{t=1}^{T} \sum_{i,j}^{N} \exp\left(-\frac{1}{2\sigma^2}\left(\lambda \cdot {d_{t}^{ij}}(t)^2 + {d_{n}^{ij}}(t)^2\right)\right)
\end{equation}

where $d_{t}^{ij}(t)$ and $d_{n}^{ij}(t)$ represent the tangential and normal distances from agent $j$'s trajectory point at time $t$ to agent $i$'s heading axis, respectively, and $\sigma$ is the standard deviation for these distances. In this formulation, $\lambda$ is a scaling factor applied to the tangential distance $d_{t}^{ij}(t)$. This approach contrasts with the disk approximation method, which primarily penalizes the Euclidean distance between agents. By accounting for both tangential and normal components, the Gaussian collision distance method significantly reduces collision rate, which we discuss in \cref{sec:experiments}.

\begin{figure}[!!t]
    \centering
    \includegraphics[width=\linewidth]{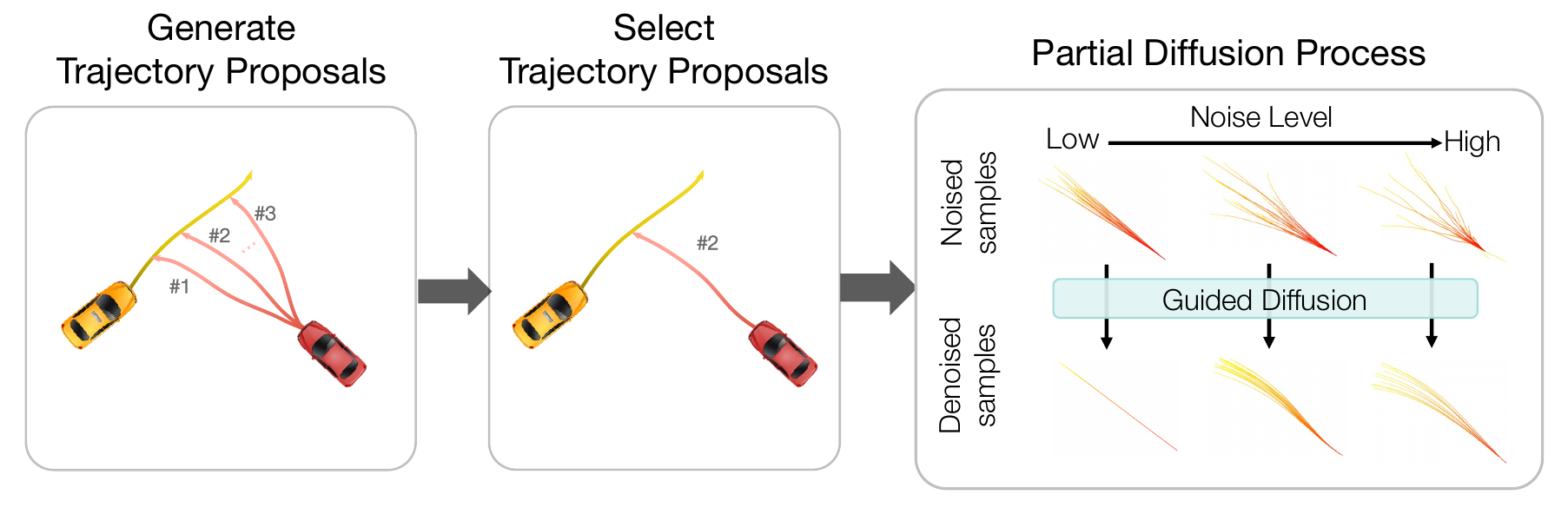}
\caption{\textbf{Framework for Partial Diffusion.} We generate proposals based on domain knowledge (e.g., collision types). Users can adjust noise levels to balance between user control and the model’s data distribution.}
    \label{fig:partial_diffusion_framework}
    \vspace{-1em}
\end{figure}

\subsection{Partial Diffusion: Controlling Collision Types}\label{sec:ctypes}

We introduce a novel approach through a partial diffusion process, utilizing trajectory proposals to initiate the diffusion process. This methodology enables the variation in collision types by the adversarial agent within the diffusion, tailoring the adversarial outcomes to specific evaluation needs, the results are discussed in \cref{sec:exp:controllable}.

Figure~\ref{fig:partial_diffusion_framework} illustrates our framework, which is divided into three main steps to generate trajectory proposals for various collision scenarios. First, we create initial trajectory proposals ($\tau_0$) aimed at capturing different types of collisions. The next critical step involves setting the partial diffusion ratio $\gamma$, which defines the specific point in the process, $k_p = \gamma \cdot K$, at which we start modifying the trajectory. Starting from step $k_p$, we adjust the trajectory by adding a precise level of Gaussian noise $\epsilon \sim N(0, I)$: $\hat{\tau}_{k_p} = \sqrt{\bar{\alpha}_{k_p}} \tau_0 + \sqrt{1-\bar{\alpha}_{k_p}} \epsilon$. The final stages include removing noise and using guided diffusion for the rest of the $k_p$ steps to refine the trajectory into a realistic path that suits our collision scenario goals.

\textcolor{black}{To generate the trajectory proposals, we develop a rule-based approach in which we first identify the centerlines of the ego and adversarial agent and then search for potential intersections of their respective centerlines. If such an intersection exists, we generate the proposals by selecting an acceleration value and lateral offset from the centerline that is likely to cause the desired collision type based on the projected plan of the ego agent. Note that the trajectory proposals are updated in a closed-loop manner to account for the interaction between the ego and the adversarial agent.
}

\textcolor{black}{
This method allows for precise control over the diffusion trajectory, enabling adversarial agents to create customized collision scenarios. Users can adjust $\gamma$ to fine-tune the balance between explicit control and the model's trained data distribution.
}

\section{Experiments}
\label{sec:experiments}

We validate the efficacy of our proposed framework via experiments with real-world driving data. Our results demonstrate that the framework can generate  \textit{realistic} and  \textit{controllable} adversarial behavior to challenge the planner. 

\subsection{Dataset}\label{sec:exp:dataset}
We conduct our experiments on two large-scale real-world driving datasets: nuScenes \cite{caesar2020nuscenes}, which  consists of 5.5 hours of driving data from two cities, and nuPlan, which consists of 1500 hours of driving data from four cities. We train the model on scenes from the nuScenes train split and evaluate it on the scenes from the nuScenes validation splits and nuPlan mini validation splits. We focus on vehicle-to-vehicle interactions.

\begin{figure}[!!t]
  \centering
  \includegraphics[width=\linewidth]{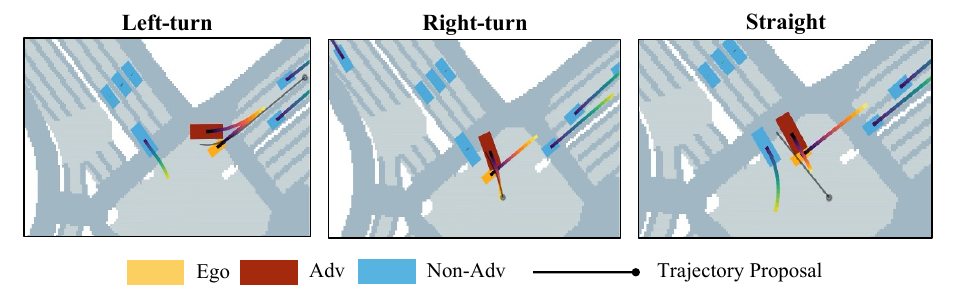}
  \caption{\textbf{Partial Diffusion Results of Rule-Based Planner on NuScenes Dataset.} The safety-critical scenarios show the framework's ability to create realistic and challenging situations, varying collision types based on different trajectory proposals. The gradient lines reflects the planned trajectories in the next 3.2 seconds.}
  \label{fig:qual}
  \vspace{-1em}
\end{figure}

\subsection{Implementation Details}\label{sec:exp_implem}

{\textbf{Baselines.} We compare our approach against STRIVE \cite{strive} using their \href{https://github.com/nv-tlabs/STRIVE}{open-source implementation} and our re-implementation of DiffScene \cite{xu2023diffscene}. STRIVE is recognized for its proficiency in generating adversarial safety-critical scenarios using a learned traffic model and adversarial optimization in the latent space.}

\textbf{Planner.} Our experiments utilize a range of different planners: 1) A rule-based planner, as implemented in STRIVE \cite{strive}, which operates on the lane graph and employs the constant velocity model to predict future trajectories of non-ego vehicles, generating multiple trajectory candidates and selecting the one least likely to result in a collision; 2) a hybrid planner BITS \cite{BITS}; 3) PDM-Closed \cite{dauner2023parting}, the winner of the 2023 nuPlan Planning Challenge; 4) a deterministic a learning-based Behavior Cloning (BC) planner, which utilizes a ResNet Encoder, followed by an MLP decoder to generate future trajectories; and 5) an Intelligent Driver Model (IDM) planner \cite{treiber2000congested}.

\textbf{Diffusion Model.} We follow the architecture described in \cite{CTG}. We represent the context using an agent-centric map and past trajectories on a rasterized map. The traffic scene is encoded using a ResNet structure \cite{he2016deep}, while the input trajectory is processed through a series of 1D temporal convolution blocks in an UNet-like architecture, as detailed in \cite{diffuser}. The model uses $K=100$ diffusion steps. During the inference phase, we generate a sample of potential future trajectories for each reactive agent in a given scene. From these, we select the trajectory that yields the lowest guidance cost. This process is referred to as \textit{filtering}.

\textbf{Closed-loop Simulation.} The simulation framework for our experiments is built upon an open-source traffic behavior framework \cite{BITS}. Within this framework, both the planner and reactive agents update their plans at a frequency of 2Hz. 

\subsection{Evaluation Metrics}\label{sec:exp:metrics}
Our goal is to validate that the proposed method can generate safety-critical scenarios that are both \textit{realistic} and \textit{controllable}. For \textit{realism} assessment, in accordance with \cite{CTG}, we compare statistical data between simulated trajectories and actual ground trajectories. This involves calculating the Wasserstein distance between their driving profiles' normalized histograms, focusing on the mean of mean values for three properties: longitudinal acceleration, latitudinal acceleration, and jerk. To evaluate \textit{controllability}, we measure metrics related to parameters we can control, specifically \textbf{relative speed} and \textbf{time-to-collision cost} between the ego and adversarial agents. 

\textcolor{black}{
Our method aims to evaluate the extent of control over adversarial behaviors within a single scenario. To do this, we focus on measuring \textit{\textbf{collision diversity}} —a metric that quantifies the range of differences in collision angles, relative speeds, and collision points. By calculating the variance of these parameters within the same scenario, we can determine how diverse and controllable the adversarial behaviors are, ensuring the scenario's realism and controllability.
}

For a more nuanced understanding of \textit{realism}, we analyze the \textbf{collision rate} – the average fractions of agents colliding, and the \textbf{offroad} rate – the percentage of reactive agents going off-road, both of which are considered failure rates \cite{BITS}. All metrics are averaged across scenarios. For detailed definitions and metrics, see supplementary \cref{sec:sup:metrics}.

\setlength\tabcolsep{8pt} 

\begin{table}[t]
\centering
\resizebox{\linewidth}{!}{%
\begin{tabular}{l|c|cccccc} 
\toprule
\textbf{Dataset}          & \textbf{Method} & \begin{tabular}[c]{@{}c@{}}\textbf{Collision} \\(\%) $\uparrow$\end{tabular} & \begin{tabular}[c]{@{}c@{}}\textbf{Other }\\\textbf{Offroad}~\\(\%)$\downarrow$\end{tabular} & \begin{tabular}[c]{@{}c@{}}\textbf{Adv }\\\textbf{Offroad} \\(\%) $\downarrow$\end{tabular} & \begin{tabular}[c]{@{}c@{}}\textbf{Collision }\\\textbf{Rel Speed} \\(m/s) $\downarrow$\end{tabular} & \begin{tabular}[c]{@{}c@{}}\textbf{Realism} \\$\downarrow$\end{tabular} & \begin{tabular}[c]{@{}c@{}}\textbf{Time} \\(s) $\downarrow$\end{tabular}  \\ 
\midrule
\multirow{3}{*}{nuScenes} & SAFE-SIM        & \textbf{43.2}                                                                & \textbf{1.8}                                                                                 & 11.4                                                                                        & \textbf{-0.12}                                                                                       & \textbf{0.38}                                                           & {$\bm{104.5 \pm 17.7}$}                                                          \\
                          & STRIVE          & 36.4                                                                         & 2.2                                                                                          & 11.4                                                                                        & 5.52                                                                                                 & 0.85                                                                    & $427.2 \pm 169.8$                                                         \\
                          & DiffScene       & 18.2                                                                         & 11.4                                                                                         & \textbf{9.0}                                                                                & 16.4                                                                                                 & 0.52                                                                    & $105.4 \pm 22.5$                                                                         \\ 
\midrule
\multirow{2}{*}{nuPlan}   & SAFE-SIM        & \textbf{80}                                                                  & \textbf{9.4}                                                                                 & 11.7                                                                                        & 6.75                                                                                                 & \textbf{0.27}                                                           & $\bm{173.4 \pm 73.3}$                                                                                                     \\
                          & DiffScene       & 56.7                                                                         & 14.0                                                                                         & \textbf{5.0}                                                                                & \textbf{-2.81}                                                                                       & 0.42                                                                    & $176.7 \pm 77.5$                                                                                                     \\
\bottomrule
\end{tabular}
}
\caption{\textbf{Safety-critical Traffic Simulation.} We compare our approach against STRIVE \cite{strive} and DiffScene \cite{xu2023diffscene} for safety-critical traffic simulation with a rule-based planner. \safesim outperforms STRIVE on all metrics and demonstrates higher collision rates and better realism than DiffScene.}
\label{tab:strive}
\end{table}

\setlength\tabcolsep{5 pt}

\begin{table}[t]
\centering
\resizebox{\linewidth}{!}{%
\begin{tabular}{l|cccccc} 
\toprule
\textbf{Planner} & \begin{tabular}[c]{@{}c@{}}\textbf{Ego-Adv Coll} \\(\%) $\uparrow$\end{tabular} & \begin{tabular}[c]{@{}c@{}}\textbf{Ego-Other Coll} \\(\%) $\uparrow$\end{tabular} & \begin{tabular}[c]{@{}c@{}}\textbf{Adv Offroad} \\(\%) $\downarrow$\end{tabular} & \begin{tabular}[c]{@{}c@{}}\textbf{Coll Speed} \\(m/s) $\downarrow$\end{tabular} & \begin{tabular}[c]{@{}c@{}}\textbf{Ego Accel } \\(m/s$^2$) $\downarrow$\end{tabular} & \begin{tabular}[c]{@{}c@{}}\textbf{Realism} \\$\downarrow$\end{tabular}  \\ 
\midrule
BC               & 38.8                                                                            & 37.3                                                                              & 9.0                                                                              & 2.47                                                                             & \textbf{0.54}                                                                        & 0.79                                                                     \\
IDM              & \textbf{49.3}                                                                   & \textbf{58.2}                                                                     & 3.0                                                                              & \textbf{-0.40}                                                                   & 0.94                                                                                 & 0.78                                                                     \\
Lane-Graph       & 34.3                                                                            & 37.3                                                                              & \textbf{1.5}                                                                     & 2.68                                                                             & 1.62                                                                                 & \textbf{0.57}                                                            \\
BITS             & 16.4                                                                            & 19.4                                                                              & 6.0                                                                              & 3.07                                                                             & 0.95                                                                                 & 0.79                                                                     \\
PDM-Closed       & 26.9                                                                            & 50.7                                                                              & \textbf{1.5}                                                                     & 2.73                                                                             & 1.30                                                                                 & 0.86                                                                     \\
\bottomrule
\end{tabular}
}
\caption{\textbf{Safety-Critical Simulation for Different Planners.} \safesim can generate diverse safety-critical scenarios tailored to different planners, including rule-based, learning-based, and hybrid planners.}
\label{tab:reb_diffplan}
\end{table}

\setlength\tabcolsep{6 pt}
\setlength\tabcolsep{4 pt}

\begin{table}[t]
\centering

\begin{tabular}{cc|cc|cc} 
\toprule
\textbf{TTC Cost} & \textbf{TTC}  & \textbf{Coll Speed} & \textbf{Coll Angle} & \textbf{Coll Rate} & \textbf{Realism}  \\
\textbf{Weight}   & \textbf{Cost} & (m/s)               & (deg)               & (\%) $\uparrow$    & $\downarrow$      \\ 
\midrule
0.0               & 0.18          & 2.45                & -7.43               & 48.2               & 0.76              \\
1.0               & 0.21          & 2.30                & 0.43                & 53.6               & 0.79              \\
2.0               & 0.26          & 3.78                & -17.0               & 60.7               & 0.81              \\
\bottomrule
\end{tabular}
\caption{\textbf{Controlling Time to Collision (TTC).} The table shows the impact of different TTC Cost weights on collision scenarios. Increasing the TTC Cost weight results in an increase in collision rate, suggesting a heightened challenge for the ego vehicle in avoiding collisions.}
\label{tab:ctrl_ttc}
\end{table}

\setlength\tabcolsep{6 pt}
\setlength\tabcolsep{2.5 pt}

\begin{table}
\centering
\resizebox{\linewidth}{!}{%
\begin{tabular}{ccc|ccc|ccc} 
\toprule
\multirow{4}{*}{$\mathbf{J_{\text{adv}}}$} & \multirow{4}{*}{$\mathbf{J_{\text{reg}}}$} & \multirowcell{4}{\textbf{Partial}\\\textbf{Diffusion}} & \multicolumn{3}{c|}{\textbf{Collision Metrics}} & \multicolumn{3}{c}{\textbf{Diversity}} \\ 
\cmidrule{4-9}
 & &  & \multirow{2}{*}{\textbf{Collision}} & \textbf{Adv} & \multirow{2}{*}{\textbf{Realism}} & \textbf{Collision} & \textbf{Collision}         & \textbf{\textbf{Collision}}   \\

 & &  & & \textbf{Offroad} &   & \textbf{Angle Var} & \textbf{Rel Speed Var}         & \textbf{Point Var}   \\

 &&                                & (\%) $\uparrow$      & (\%) $\downarrow$    & $\downarrow$         & (rad) $\uparrow$        & (m/s) $\uparrow$ & (m) $\uparrow$  \\ 
\midrule
\textbf{\textcolor{Green}{\checkmark}} & \textbf{\textcolor{Green}{\checkmark}} & \textcolor{red}{$\mathbf{\times}$}  & 23.9                 & \textbf{13.8}        & 0.58                & 2.22                    & \multicolumn{1}{c}{2.99}            & 1.62        \\
\textbf{\textcolor{Green}{\checkmark}} & \textbf{\textcolor{Green}{\checkmark}}  & \textbf{\textcolor{Green}{\checkmark}}   & 29.0                 & 14.6                 & \textbf{0.57}                 & 3.10                    & \multicolumn{1}{c}{1.96}            & \textbf{5.44}           \\
\textbf{\textcolor{Green}{\checkmark}} & \textcolor{red}{$\mathbf{\times}$} & \textcolor{red}{$\mathbf{\times}$} & \textbf{53.5}                 & 23.1                 & 0.58                 & \textbf{3.34}                    & \textbf{4.81}            & 2.47           \\
\bottomrule
\end{tabular}
}

\caption{\textbf{Ablation Study on Controllability.} This study examines each component of our proposed method. Partial diffusion significantly increases the variance of the collision point, resulting in a greater diversity of collision scenarios. }
\label{tab:coll_diversity}
\end{table}

\setlength\tabcolsep{6 pt}


\subsection{Evaluation of Safety-Critical Traffic Simulation with Baseline Methods}\label{sec:exp:safety_critical}

We compared our method with STRIVE \cite{strive} and DiffScene\cite{xu2023diffscene}, utilizing the Lane-graph-based planner from STRIVE. The evaluation focused on collision rates between the ego and the adversarial agent (“Collision”), adversarial agent off-road rates (“Adv Offroad”), other agents’ off-road rates (“Other Offroad”), collision relative speed between the ego and adversary (“Collision Rel Speed”), realism of all non-ego agents (“Realism”), and simulation time.

The results, presented in \cref{tab:strive}, demonstrate that our method excels in all metrics compared to STRIVE, especially in collisions and realism. Compared to DiffScene, \safesim also exhibits a higher collision rate with better realism. Note that we trained models on nuScenes and tested them in nuPlan without additional fine-tuning. As illustrated in Figure \ref{fig:qual}, the qualitative examples from the NuScenes dataset demonstrate how our framework can challenge the rule-based planner with various driving situations. See supplementary \cref{supp:qual} for more qualitative examples.

\subsection{Safety-Critical Simulation with Different Planners}\label{sec:exp:diff_planners}

In \cref{tab:reb_diffplan}, we demonstrate our framework's ability to generate collisions across various planner types: rule-based, learning-based, and hybrid planners. Notably, the IDM planner exhibits the highest Ego-Adv collision rate. This heightened rate can be attributed to the IDM's focus on vehicles near the same lane, potentially overlooking other vehicles in the scene. Consequently, in scenarios with the IDM planner, our framework can induce collisions at relatively lower speeds.




\subsection{Evaluation: Controlling Safety-Criticality}\label{sec:exp:controllable}

A key feature of \safesim is its ability to generate controllable adversarial behaviors, offering variations not possible with previous methods.

\textbf{Controlling Time-to-Collision.} We control the orientation and relative speed together using the time-to-collision (TTC) cost, as described in \cref{sec:guided_advsec}. We manipulate the scenario's safety-criticality by adjusting the relative weight of the TTC cost. To assess the impact of these adjustments, we measure the average TTC cost shortly before a collision occurs (0.5 seconds). Our observations, detailed in  \cref{tab:ctrl_ttc}, show that increasing the TTC weight raises the TTC cost. Notably, while the relative collision speed remains fairly consistent, the collision angle shifts, indicating a greater difficulty in avoiding ego-adversary collisions. Additionally, our method can control other aspects, such as relative speed, as detailed in supplementary \cref{supp:ctrl_v}.

\subsection{Ablation Study on Controllability}\label{sec:exp:pd_ablation}
We performed an ablation study to assess the influence of different guidance strategies in diffusion models on the quality of simulations. This study, detailed in Table~\ref{tab:coll_diversity}, aimed to quantify collision diversity. We compare against our baseline approach, consisting of regularized and adversarial objectives ($J_{reg} + J_{adv}$), by predetermining the selection of adversarial agents and conducting experiments with multiple seeds (three) in our framework to evaluate collision diversity. In partial diffusion, we manipulated the trajectory proposal selection mechanism across centerlines with normal offsets of -2.0, 0.0, and 2.0.

\textbf{Effectiveness of Partial Diffusion.}
Table~\ref{tab:coll_diversity} reveals that partial diffusion significantly enhances both the collision point and angle diversity in comparison to the baseline approach. This underscores the capability of partial diffusion to generate a wider array of collision scenarios through various trajectory proposals, illustrating its potential to explore diverse collision dynamics effectively.

\textbf{Impact of the Regularization Term ($J_{reg}$).}
Incorporating $J_{reg}$ results in a notable decrease in the adversarial-collision rate, highlighting the importance of the regularization term in enhancing simulation realism. Without $J_{reg}$, the collision rate between the adversarial agent and ego vehicle increases, but the adversarial vehicle goes offroad more often, leading to scenarios that differ significantly from realistic behavior.

\subsection{Limitation and Failure Cases}
We identified areas for improvement in \safesim\ (see supp. \cref{fig:failure_cases}). In certain cases, the adversarial agent unrealistically collides with non-adversarial agents before reaching the ego agent. Additionally, some scenarios result in collisions where the ego planner is not at fault. While understanding how the ego planner can avoid such cases is important, creating more scenarios where the ego is at fault would be beneficial.

\section{Conclusion}
\label{sec:conclusion}

In this work, we present a closed-loop simulation framework utilizing guided diffusion models for creating safety-critical scenarios to assess autonomous vehicle (AV) algorithms. Our research is in line with the goals of SO-TIF, focusing on how autonomous vehicles respond to dynamic scenarios like aggressive driving, underscoring our dedication to safety across diverse and unforeseeable conditions. Our framework introduces innovative guidance objectives tailored for controllable, stable, long-term safety-critical simulations. A key aspect of our method lies in its ability to vary the types of adversarial behavior within collision scenarios. By integrating adversarial objectives and partial diffusion, we enable fine-grained control over adversarial actions. This versatility enables our framework to produce a broader range of realistic and manageable scenarios, setting a new standard in adversarial scenario generation beyond the limitations of existing approaches.

Future directions for our research include: 1) exploring the application of our framework in closed-loop policy training, and 2) developing automated methods for adjusting controllable parameters. These methods aim to facilitate the generation of diverse, long-tail scenarios.  We believe this framework holds significant promise for enhancing real-world AV safety.

\setcounter{secnumdepth}{0}
\section{Acknowledgements}
\setcounter{secnumdepth}{2}
This work was part of W.J. Chang’s summer internship at NEC Labs America, and he is also supported by the National Science Foundation Graduate Research Fellowship Program under Grant No. DGE 2146752. Any opinions, findings, and conclusions or recommendations expressed in this material are those of the author(s) and do not necessarily reflect the views of the National Science Foundation. The authors would like to thank Chih-Ling Chang for her insightful suggestions and assistance with figures and presentations.

%
%

\bibliographystyle{splncs04}
\bibliography{main}

\clearpage

\setcounter{page}{1}
\setcounter{section}{0}
\setcounter{figure}{0}
\setcounter{table}{0}
\setcounter{equation}{0}
\setcounter{footnote}{0}
\renewcommand{\thepage}{A\arabic{page}}
\renewcommand{\thesection}{\Alph{section}}
\renewcommand{\thefigure}{A\arabic{figure}}
\renewcommand{\thetable}{A\arabic{table}}
\renewcommand{\theequation}{A\arabic{equation}}

\def\httilde{\mbox{\tt\raisebox{-.5ex}{\symbol{126}}}}

\makeatletter
\def\@thanks{}
\makeatother

\pagenumbering{gobble}

\title{Supplementary Material:  \safesim: Safety-Critical Closed-Loop Traffic Simulation with Diffusion-Controllable Adversaries}

\titlerunning{Safety-Critical Closed-Loop Traffic Simulation with Controllable Adversaries}



\author{ Wei-Jer Chang\inst{1} \and
Francesco Pittaluga\inst{2} \and
Masayoshi Tomizuka\inst{1} \and
Wei Zhan\inst{1} \and
Manmohan Chandraker\inst{2,3}
}


\authorrunning{W.J. Chang et al.}

\institute{$^1$ UC Berkeley \hspace{0.5cm} $^2$ NEC Labs America \hspace{0.5cm} $^3$ UC San Diego}

\maketitle

\begin{figure*}[!t]
    \centering
    \includegraphics[width=1\textwidth]{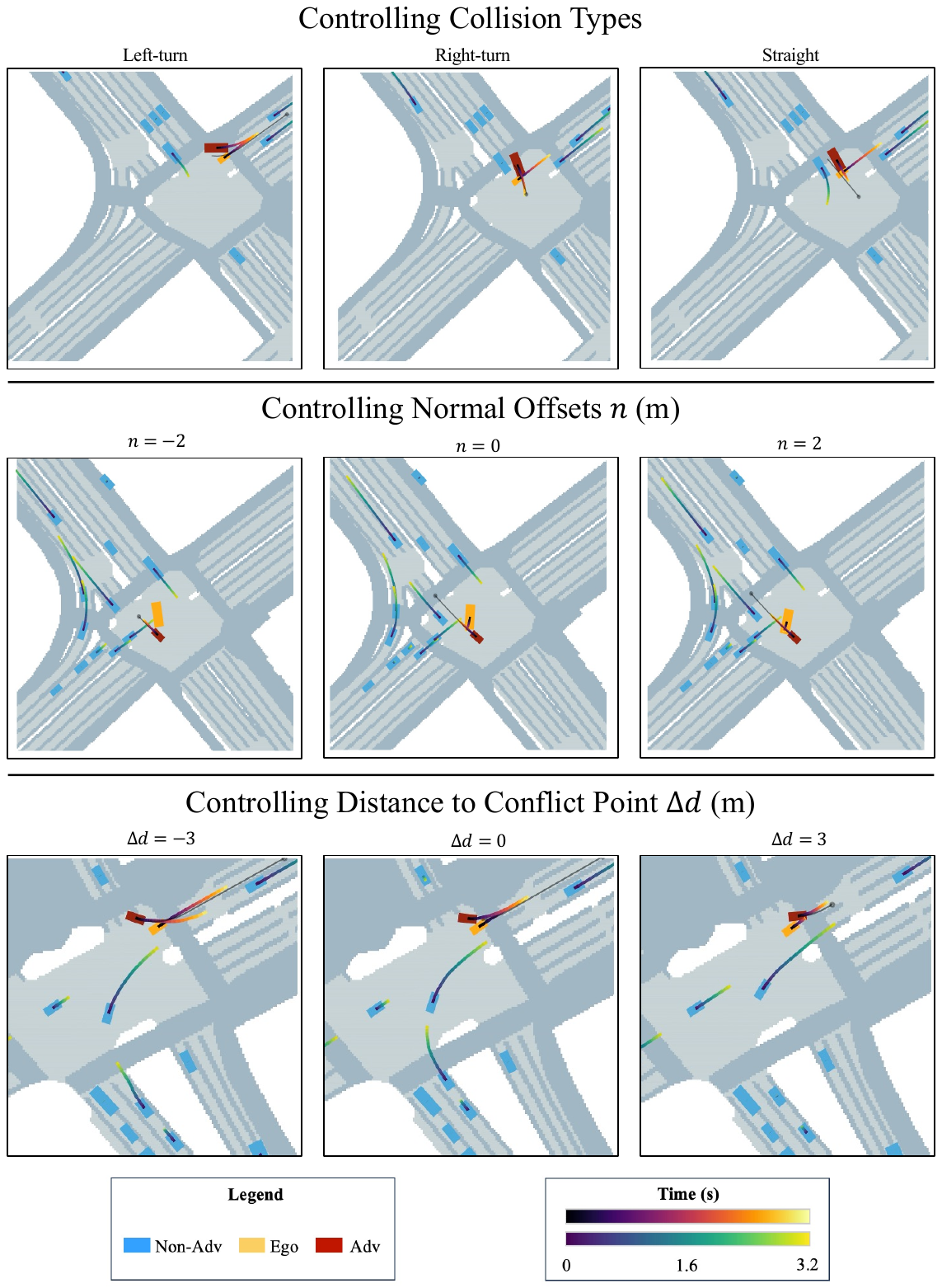}
    \caption{\textbf{Illustration of Diverse Collision Scenarios via Partial Diffusion.} This figure showcases example simulations that highlight how varying trajectory proposals can influence the occurrence and type of collisions. The black line represents the trajectory proposals for the adversarial vehicle.}
    \label{fig:partial_diffusion}
\end{figure*}

\begin{figure*}[!t]
    \centering
    \includegraphics[width=1\textwidth]{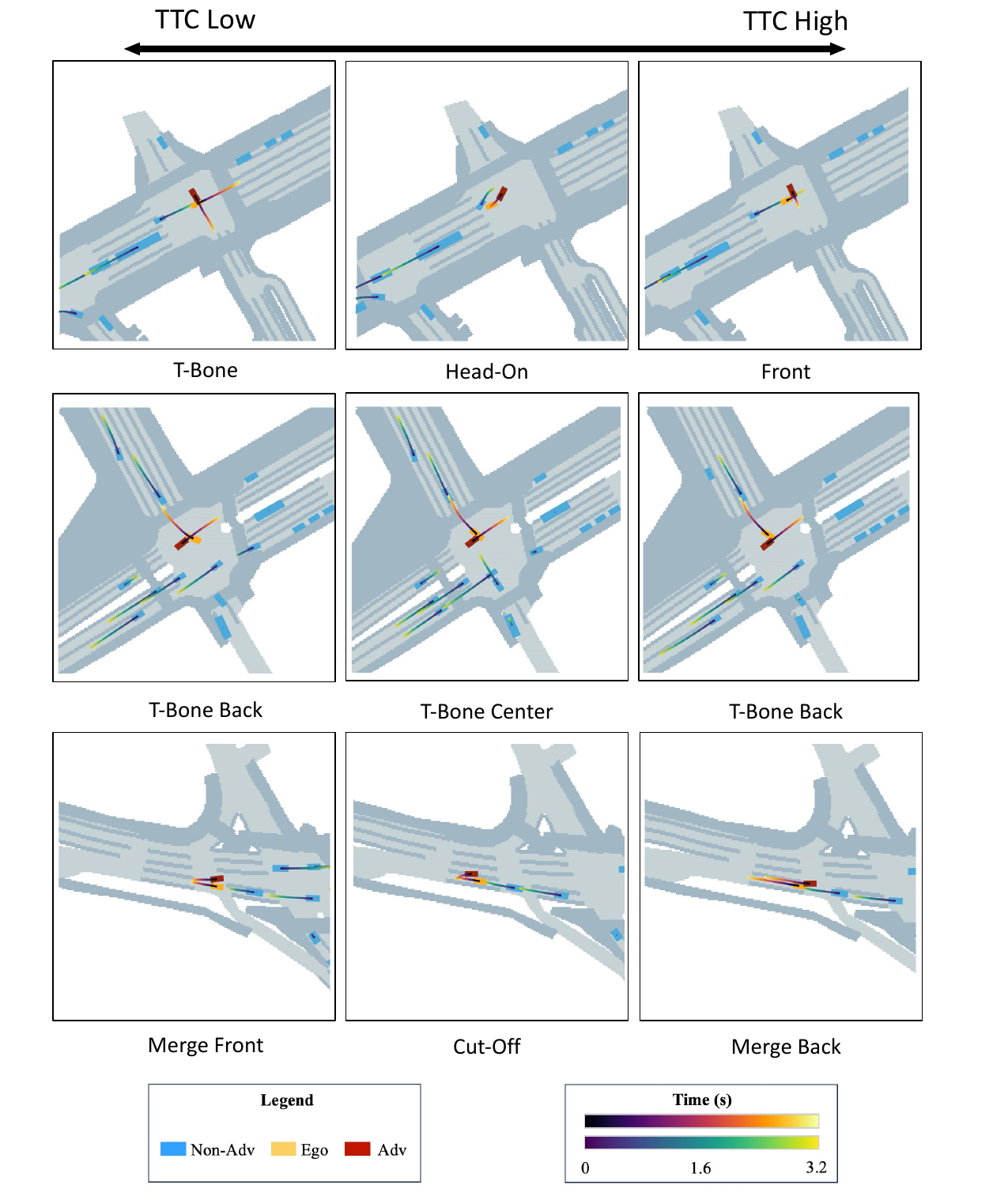}
    \caption{\textbf{Impact of Time-To-Collision (TTC) Control on Collision Scenarios.} This figure demonstrates example simulations where adjusting the TTC parameter influences the dynamics and outcomes of collision scenarios, showcasing the method's versatility in testing autonomous driving algorithms under different conditions.}

    \label{fig:ttc}
\end{figure*}

Compared to previous works, our methodology enables controllable adversaries through multiple controllable factors to generate closed-loop safety-critical simulations. This allows for the generation of a broad range of safety-critical behaviors across diverse scenarios.
\section{Qualitative Results}\label{supp:qual}

For insights into closed-loop simulation outcomes, we invite readers to view the supplementary videos.

We present two sets of qualitative results. The first set, illustrated in Figure \ref{fig:partial_diffusion}, displays a variety of safety-critical simulations where altering the trajectory proposals modifies the collision types. The second set, depicted in Figure \ref{fig:ttc}, showcases simulations that demonstrate different collision scenarios achieved by adjusting the Time-To-Collision (TTC) to influence the safety-criticality of the situation. Unlike the STRIVE method, which tends to generate scenarios with limited variability, our approach utilizes multiple control mechanisms (such as varying trajectory proposals and safety-criticality levels) to create a broader spectrum of safety-critical conditions. This flexibility is particularly beneficial for testing and evaluating autonomous driving algorithms under various challenging conditions.

\section{Details on Partial Diffusion}

\subsection{Methodology for Generating Trajectory Proposals}

\begin{figure*}[!t]
    \centering
    \includegraphics[width=1\textwidth]{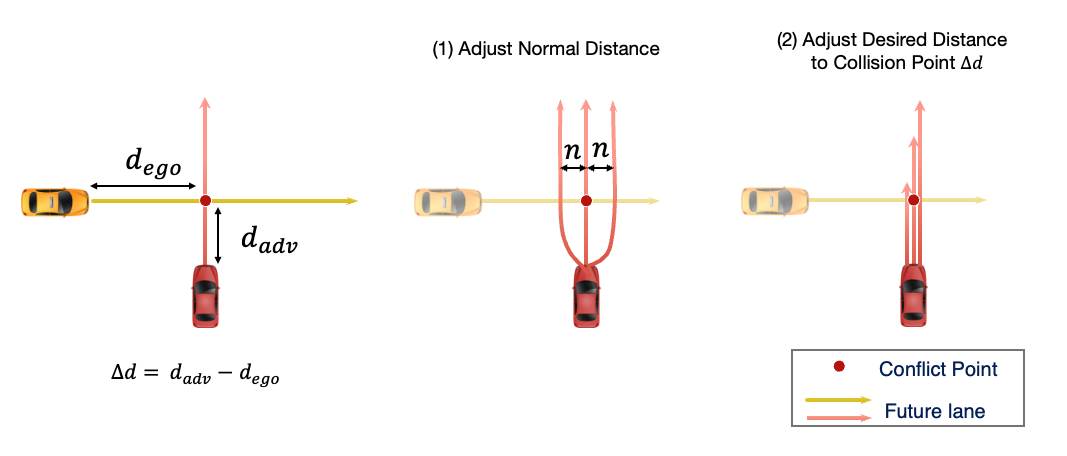}
    \caption{\textbf{Methods for generating different trajectory proposals.}}
    \label{fig:trajectory_proposals}
\end{figure*}

To generate trajectory proposals for the partial diffusion process, which aims to create potential collision scenarios, we present a straightforward method based on lane relationships. In addition to selecting different lane relationships to represent various types of collisions, we further refine our control over these scenarios by introducing two primary variations: 1) the relative distance to the conflict point and 2) the normal offsets of the lane,  as illustrated in Figure \ref{fig:trajectory_proposals}:

\begin{enumerate}
    \item \textbf{Relative Distance to the Conflict Point:} This adjustment allows for the precise management of how vehicles navigate interactions, such as passing or yielding, by selecting specific accelerations that achieve the desired distance to the conflict point.
    \item \textbf{Lane's Normal Offsets:} Modifying these offsets enables the generation of trajectories that accurately reflect the spatial dynamics of vehicle positioning within lanes.
\end{enumerate}
In addition, we can also generate proposals based on different lanes to have different relationships. 

Note that is essential to generate trajectory proposals within a closed-loop simulation, updated at every planning cycle. Since the diffusion model outputs action sequences, after generating the initial proposals, we employ inverse dynamics to calculate the corresponding turning rates.

\subsection{Ablation study of Partial Diffusion}
We measure the Mean Squared Error (MSE) to quantify the difference between initial trajectory proposals and the outcomes from the partial diffusion model, focusing on the first second of the trajectory in each planning iteration. Table \ref{tab:partial_ablation} reveals that the trajectory MSE varies with the diffusion ratio. This ratio is adjustable, enabling the calibration of the model to align with user needs and maintain a balance between the original proposals and the diffusion model's output. A partial diffusion ratio of $\gamma = 0.0$ corresponds to the highest collision rate, suggesting that our initial trajectory proposals effectively signal potential collisions. After the diffusion model's denoising step, the lateral acceleration diminishes significantly, leading to more realistic trajectory generations. This underscores the importance of our proposed partial diffusion process, highlighting its effectiveness in balancing the alignment between trajectory proposals and the model's output, which represents the underlying data distribution.

\setlength\tabcolsep{4pt} 

\begin{table}
\centering

\label{tab:strive2}
\begin{tabular}{ccccc} 
\toprule
\multicolumn{1}{l}{\multirow{2}{*}{\begin{tabular}[c]{@{}l@{}}\textbf{Partial Diffusion }\\\textbf{ratio~$\gamma$}\end{tabular}}} & \textbf{Coll Rate} & \textbf{Adv Offroad} & \textbf{Traj MSE}    & \begin{tabular}[c]{@{}c@{}}\textbf{Adv max }\\\textbf{lateral acc}\end{tabular}  \\
\multicolumn{1}{l}{}                                                                                                              & (\%) $\uparrow$    & (\%) $\downarrow$    & ($m^2$) $\downarrow$ & ($m/s^3$)~$\downarrow$                                                           \\ 
\midrule
0.0                                                                                                                               & \textbf{26.8}      & \textbf{7.5}         & \textbf{3.09}        & 17.5                                                                             \\
0.2                                                                                                                               & 14.6               & 12.5                 & 20.2                 & 2.3                                                                              \\
0.4                                                                                                                               & 17.1               & 10.0                 & 19.2                 & 2.3                                                                              \\
0.6                                                                                                                               & 9.8                & 12.5                 & 16.6                 & 2.0                                                                              \\
0.8                                                                                                                               & 12.2               & 10.0                 & 22.2                 & 2.44                                                                             \\
1.0                                                                                                                               & 14.6               & \textbf{7.5}         & 35.6                 & \textbf{1.75}                                                                    \\
\multicolumn{1}{l}{w/o Partial Diffusion}                                                                                         & 7.4                & 10.0                 & 33.4                 & 2.22                                                                             \\
\bottomrule
\end{tabular}
\vspace{.2em} 
\caption{\textbf{Ablation study on the Partial Diffusion ratio.}}
\label{tab:partial_ablation}
\vspace{.2em} 
\setlength\tabcolsep{6pt} 
\end{table}

\section{Metrics Definitions}\label{sec:sup:metrics}
This section outlines the definitions of the metrics used in our evaluations, averaged across all scenarios, except for the realism metric.

\subsection{Traffic Simulation Metrics}

\paragraph{Off-road.} This metric measures the percentage of agents that go off-road in a given scenario. An agent is considered off-road if its centroid moves into a non-drivable area.

\paragraph{Collision.} This metric represents the percentage of agents involved in collisions with other agents during the simulation. 


\paragraph{Realism.} Adopting the approach from \cite{CTG}, realism is quantified using the Wasserstein distance. This metric compares the normalized histograms of the driving profiles, focusing on the mean values of three key properties: longitudinal acceleration, lateral acceleration, and jerk. A lower value indicates a higher degree of realism.


\subsection{ Adversarial Behavior and Collision Metrics}\label{supp:adv_behavior_def}

\paragraph{Collision Relative Speed.} Collision Relative Speed is defined as the ego planner's speed minus the adversarial vehicle's speed at the collision timestep. 

To control the relative speed, we introduce the relative speed cost function:
\begin{equation}
J_v = \sum_{t=1}^{T} |v^1_t - v^a_t - v_{\text{diff}}| \cdot \mathbf{1}{\{d(t) < d_{\text{col}}\}},
\end{equation}
where $v_{\text{diff}}$ is the desired speed difference between the ego and the adversarial vehicles, influencing the relative speed at the point of collision. The function $\mathbf{1}{\{d(t) < d_{\text{col}}\}}$ is an indicator function that applies the cost only when the distance $d(t)$ between the ego and adversarial vehicle is less than a specified threshold $d_{\text{col}}$.

\paragraph{Time-to-Collision Cost.} The Time to Collision (TTC) cost \cite{nishimura2023rap} assesses collision risk based on the relative speed and orientation between agents. For two agents located at positions \((x_i, y_i)\) and \((x_j, y_j)\) with respective velocities \((v_{x_i}, v_{y_i})\) and \((v_{x_j}, v_{y_j})\), we define their relative position and velocity. The relative position is given by \( dx = x_i - x_j \) and \( dy = y_i - y_j \), representing the positional differences along the x and y axes. Similarly, the relative velocity is calculated as \( dv_x = v_{x_i} - v_{x_j} \) and \( dv_y = v_{y_i} - v_{y_j} \), which are the differences in their velocities along the x and y axes.
 The TTC is computed under a constant velocity assumption, solving a quadratic equation to find the time of collision \(t_{\text{col}}\), with a collision considered when relative distance is minimal.

The real part of the solution provides the time to the point of closest approach, \( \tilde{t}_{\text{col}} \), calculated as:
\begin{equation}
\tilde{t}_{\text{col}} = \begin{cases} 
- \frac{dv_x dx + dv_y dy}{\tilde{dv}^2} & \text{if } \tilde{t}_{\text{col}} \geq 0, \\
0 & \text{otherwise},
\end{cases}
\end{equation}
and the distance at that time, \( \tilde{d}_{\text{col}} \), is given by:
\begin{equation}
\tilde{d}_{\text{col}}^2 = \begin{cases} 
\frac{(dv_x dy - dv_y dx)^2}{\tilde{dv}^2} & \text{if } \tilde{t}_{\text{col}} \geq 0, \\
dx^2 + dy^2 & \text{otherwise}.
\end{cases}
\end{equation}

 We define the TTC cost $J_{\text{ttc}}$ as:
\begin{equation}
J_{\text{ttc}} = \sum_{t=1}^{T} - \exp\left( -\frac{\tilde{t}_{\text{col}(t)}^2}{2\lambda_t} - \frac{\tilde{d}_{\text{col}(t)}^2}{2\lambda_d} \right),
\end{equation}
where \(\lambda_t\) and \(\lambda_d\) are the time and distance bandwidth parameters. This cost is evaluated over a time horizon \(T\), with a higher cost for scenarios having low time to collision and proximity. For further details on the derivation of this cost function, we direct readers to \cite{nishimura2023rap}.

In our evaluations, we focus on the average TTC cost of 0.5 seconds preceding a collision. This metric effectively captures the criticality of the safety scenarios, reflecting the potential risk of imminent collisions.

\paragraph{Time-to-Collision.}Additionally, we compute the average Time-to-Collision (TTC) for each timestep within the crucial 0.5-second window before collisions occur in our scenarios. It's important to note that this TTC is not the actual time until a collision, but rather a theoretical estimate based on the constant velocity model assumption for each timestep.

\section{Implementation Details}\label{sec:supp_implmentation_details}
In this section, we discuss the implementation details of our diffusion model and the experimental settings.

\subsection{Diffusion Model Training and Parameterization}
The training objective is to minimize the expected difference between the true initial trajectory and the one estimated by the model, formalized by the loss function \cite{diffusion_clean}\cite{CTG}:
\begin{equation}
\mathcal{L} = \mathbb{E}_{\epsilon, k, \tau_0, c} \left[ \|\tau_0 - \hat{\tau}_0\|^2 \right]
\end{equation}
where $\tau_0$ and $\mathbf{c}$ are sampled from the training dataset, $k \sim \mathcal{U}\{1, 2, \ldots, K\} $ is the timestep index sampled uniformly at random, and  $ \epsilon \sim \mathcal{N}(0, \mathbf{I})$ is Gaussian noise used to perturb $\tau_0$ to produce the noised trajectory $\tau_k$.

In each denoising step, our model predicts the mean of the next denoised action trajectory \cref{eq:3}. Instead of predicting the noise $\epsilon$ that is used to corrupt the trajectory \cite{MID}, we directly output the denoised clean trajectory $\hat{\tau}_0$ \cite{diffusion_clean}\cite{CTG}. The predicted mean based on  $\hat{\tau}_0$ and  ${\tau}_k$:
\begin{equation}
\tau_{k-1} = \mu_{\theta}(\tau_k, \hat{\tau}_0) = \frac{\sqrt{\bar{\alpha}_{k-1}} \beta_k}{1 - \bar{\alpha}_k} \hat{\tau}_0 + \frac{\sqrt{\alpha_k} (1 - \bar{\alpha}_{k-1})}{1 - \bar{\alpha}_k} \tau_k
\end{equation}
where $ \beta_k $ represents the variance from the noise schedule in the diffusion process, $ \alpha_k $ is defined as $ \alpha_k := 1 - \beta_k $, indicating the incremental noise reduction at each step, and $ \bar{\alpha}_k $ is the cumulative product of $ \alpha_j $ up to step $ k $, mathematically expressed as $\bar{\alpha}_k = \prod_{j=0}^k \alpha_j$.
\subsection{Diffusion Process Details}
For the diffusion process, we utilize a cosine variance schedule as described in \cite{diffuser}, with the number of diffusion steps set to $K = 100$. The variance scheduler parameters are configured with a lower bound $\beta_1$ of 0.0001 and an upper bound $\beta_K$ of 0.05. The diffusion model takes in a 1-second history and is trained to predict the next 3.2 seconds with a step time $dt=0.1$. Our model was trained on four NVIDIA RTX A6000 GPUs for 70000 iterations using the Adam optimizer, with a learning rate set to $1 \times 10^{-5}$. The diffusion model's implementation is based on methodologies from open-source repositories \cite{diffuser, MID}, and the simulation framework is developed based on \cite{ivanovic2023trajdata, BITS}.


\subsection{Guidance details} 
To simultaneously incorporate multiple guidance functions in our model, we assign weights to balance their contributions. In our experiments, particularly with non-adversarial agents, we implement a combination of route guidance ($J_{\text{route}}$) and Gaussian collision guidance ($J_{\text{gc}}$) across $M=20$ examples. Notably, we apply a filtration process exclusively to $J_{\text{gc}}$, aiming to prevent imminent collisions For adversarial agents, we maintain the same weighting across all guidance functions, but uniquely control the weighting for $J_{\text{ttc}}$ to achieve controllable behavior. In this setting, we select the sample that yields the highest adversarial cost ($J_{\text{adv}}$), ensuring effective and targeted adversarial scenarios.

\textbf{Collision Guidance}: The collision guidance is based on different agent interactions. Following the methodology of \cite{CTG}, we extend the denoising process of all agents within a scene into the batch dimension. During inference, to generate $M$ samples, we proceed under the assumption that each sample corresponds to the same $m$-th example of the scene. For the ego vehicle, the future state predictions are derived from a diffusion model identical to the one used for other agents. The collision distance for the ego vehicle is then computed considering these predictions and their interactions with other agents within the scene.

\subsection{Selecting Adversarial Agents}\label{supp:selecting_adv}
\textcolor{black}{ To effectively select adversarial agents for safety-critical simulation, we developed two strategies: dynamically selecting adversarial agents or selecting interacting agents. Inspired by \cite{strive}, we proposed to dynamically adjusting the weighting coefficient $\rho^i$ of $J_{\text{adv}}$ during the guided diffusion process, encouraging a collision by minimizing the positional distance between controlled agents and the tested ego car:
\begin{equation}\label{eq:select_adv}
\rho_{i,t} = \frac{\exp(-d^{i,1}(t))}{\sum_{j} \exp(-d^{j,1}(t))}
\end{equation}
where $d^{i,1}(t)$ represents the euclidean distance between agent $i$ and the ego vehicle at time $t$. Intuitively, the $\rho^{i,t}$ coefficients, defined by the softmax operation, identify a candidate agent to collide with the ego vehicle. The agent with the highest $\rho^{i,t}$ value is considered the most likely “adversary” based on proximity, and this formulation prioritizes causing a collision with this adversary. This approach weights the adversarial loss $J_{\text{adv}}$ to highlight key interactions, preventing the unrealistic of all agents acting adversarially towards the ego vehicle.}

\textcolor{black}{ An alternate strategy selects interacting agents as adversaries based on their lane positions relative to the ego. Agents within a certain lane proximity to the ego are randomly chosen. In this scenario, the selected $i$th agent is treated as $\rho^{i,t} = 1$, with all others set to zero, for the duration of the simulation.}

\section{Experimental Settings.}
We dynamically select adversarial agents as described in \cref{eq:select_adv}, based on the criteria outlined in \cref{tab:strive}. In contrast, Tables 3, 4, and 5 use preselected and fixed adversarial agents. Additionally, Tables 3 and 4 focus on intersection scenarios where interactions are more involved. The selected scenarios will be available at our \href{https://safe-sim.github.io/}{webpage}.

\begin{table}[t]
\centering
\begin{tabular}{c|c|cc}
\toprule
 \textbf{Rel Speed Control} & \textbf{Ego-Adv Rel Speed} & \textbf{Coll Rate} & \textbf{Realism}  \\ 
   (m/s) & (m/s) & (\%) $\uparrow$ & $\downarrow$ \\ 
\midrule
-2.0  & 0.90    & 0.29   & 0.83      \\
0.0       & 1.26    & 0.38   & 0.89      \\
2.0   & 1.94    & 0.44   & 0.88      \\ 
\bottomrule
\end{tabular}
\caption{\textbf{Controlling relative collision speed.} This table illustrates the ability of our framework to modulate the relative speed between ego and adversarial agents, influencing collision rates while maintaining realism. }

\label{tab:ctrl_v}
\end{table}

\section{Additional Experiments}
\subsection{Controllability: Controlling Relative Speed.}\label{supp:ctrl_v}
In our safety-critical simulation framework, we examine the effects of manipulating the desired relative speed between the ego vehicle and the adversarial agent. As shown in ~\cref{tab:ctrl_v}, our proposed relative speed control results in a notable impact on both the actual ego-adversary relative speed and the collision rate. For instance, setting a lower desired relative speed target (-2.0 m/s) generally results in a decreased ego-adversary relative speed, and vice versa for a higher target (2.0 m/s). However, these adjustments do not directly translate to matching values in the simulations due to the nature of closed-loop interactions. The planner's reactive behavior to the adversarial agent's actions contributes to this discrepancy, as it may take evasive maneuvers or adjust its speed, potentially avoiding collisions altogether. Moreover, the realism metric across different relative speed settings remains relatively consistent, suggesting that the adjustments do not compromise the realism of the driving scenarios. 
\begin{table}[t]
\centering
\resizebox{\linewidth}{!}{%
\begin{tabular}{l|cccccc} 
\toprule
\textbf{Method} & \textbf{Collision} & \textbf{Other Offroad} & \textbf{Other Collision}    & \textbf{Adv Offroad}      & \textbf{Collision Rel Speed} & \textbf{Realism}  \\
& (\%) $\uparrow$ & (\%) $\downarrow$ & (\%) $\downarrow$ & (\%) $\downarrow$ & (m/s) $\downarrow$ & $\downarrow$                       \\ 
\cmidrule{1-7}
Ours                             & \textbf{43.2}                       & \textbf{1.9}      & 1.90                                  & \textbf{11.4}     & \textbf{-0.12}     & 0.38                               \\
Our (-$J_{route}$)                 & 38.6                                & 5.6               & 2.91                                  & 15.9              & 1.07               & \textbf{0.29}                      \\
Ours (-$J_{col}$)                 & 25.0                                & 4.9               & \textbf{1.41}                         & \textbf{11.4}     & 0.94               & 0.33                               \\
\bottomrule
\end{tabular}
} 

\caption{\textbf{Ablation Study for $J_{reg}$.}}
\label{tab:reb_strive_abl}
\end{table}

\begin{figure}[!t]
    \centering
    \includegraphics[width=\linewidth]{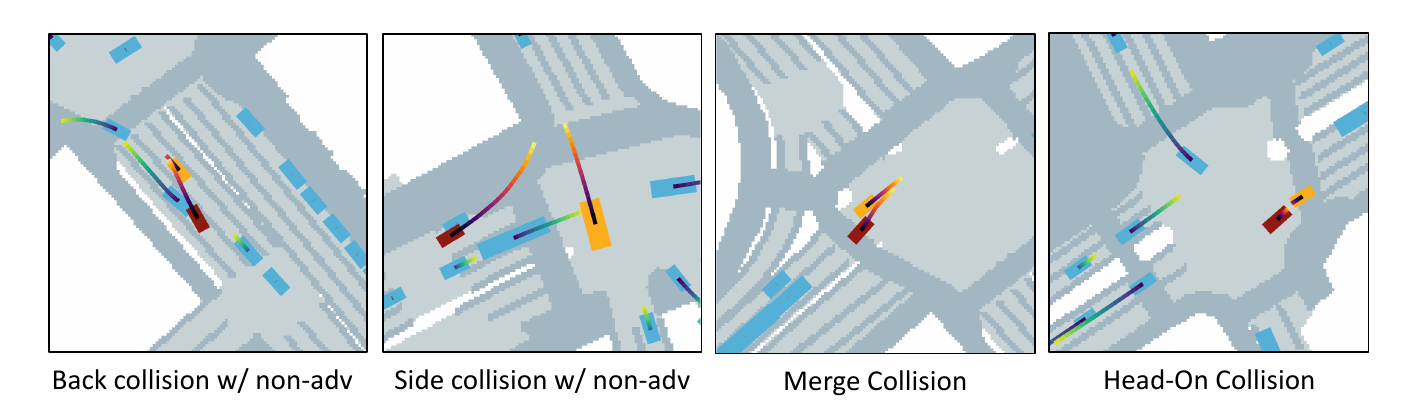}
    \caption{\textbf{Qualitative Samples of \safesim Limitation and Failure Cases.} In certain scenarios, the adversarial agent collides with non-adversarial agents before challenging the ego agent. Additionally, the adversarial agent may cause at-fault collisions.}
    \label{fig:failure_cases}
\end{figure}
\subsection{Ablation Study for $J_{reg}$.}\label{supp:j_reg}

We provide ablation study for the regularization term $J_{reg}$. Note, for \cref{tab:coll_diversity} in the main paper, adversarial agents were selected before simulation based on their lane proximity to the ego. For \cref{tab:reb_strive_abl}, adversarial agents were selected dynamically during simulation via \cref{eq:select_adv}. 

\subsection{Qualitative Analysis of \safesim's Limitations}
\label{sec:qualitative-results}
In \cref{fig:failure_cases}, we present qualitative examples highlighting areas where \safesim can be improved, including collisions with non-adv agents and at-fault collisions.

\end{document}